\PassOptionsToPackage{prologue,dvipsnames,table}{xcolor}
\documentclass[letterpaper]{article} 
\usepackage{aaai25}  
\usepackage{times}  
\usepackage{helvet}  
\usepackage{courier}  
\usepackage[hyphens]{url}  
\usepackage{graphicx} 
\usepackage{natbib}  
\usepackage{caption} 
\usepackage{amsmath}
\usepackage{booktabs}
\usepackage{multirow}
\usepackage{tcolorbox}
\usepackage[table]{xcolor}
\usepackage{arydshln}

\usepackage{algorithm}
\usepackage{algorithmic}
\usepackage{pifont}
\usepackage{makecell}
\usepackage{newfloat}
\usepackage{listings}
\DeclareCaptionStyle{ruled}{labelfont=normalfont,labelsep=colon,strut=off} 
\floatstyle{ruled}
\newfloat{listing}{tb}{lst}{}
\floatname{listing}{Listing}
\title{Neural-Symbolic Collaborative Distillation: \\ Advancing Small Language Models for Complex Reasoning Tasks}
\author{Huanxuan Liao$^{1,2}$, Shizhu He$^{1,2}$\thanks{Corresponding author}, Yao Xu$^{1,2}$, Yuanzhe Zhang$^{4}$, Kang Liu$^{1,2,3}$, Jun Zhao$^{1,2}$}
\affiliations{
    \textsuperscript{\rm 1} The Key Laboratory of Cognition and Decision Intelligence for Complex Systems, \\
    Institute of Automation, Chinese Academy of Sciences, Beijing, China \\
    \textsuperscript{\rm 2} School of Artificial Intelligence, University of Chinese Academy of Sciences, Beijing, China \\
    $^3$ Shanghai Artificial Intelligence Laboratory, Shanghai, China \\
    $^4$ National Science Library, Chinese Academy of Sciences, Beijing, China\\
    {liaohuanxuan2023@ia.ac.cn} {\{yao.xu, shizhu.he, kliu, jzhao\}@nlpr.ia.ac.cn} \\
}

\usepackage{bibentry}

\begin{document}

\maketitle

\begin{abstract}
In this paper, we propose \textbf{Ne}ural-\textbf{Sy}mbolic \textbf{C}ollaborative \textbf{D}istillation (\textbf{NesyCD}), a novel knowledge distillation method for learning the complex reasoning abilities of Large Language Models (LLMs, e.g., \textgreater 13B). We argue that complex reasoning tasks are difficult for Small Language Models (SLMs, e.g., $\leq$ 7B), as these tasks demand not only general cognitive abilities but also specialized knowledge, which is often sparse and difficult for these neural-based SLMs to effectively capture. Therefore, NesyCD distills the general capabilities and specialized knowledge in LLMs in different ways.
On the one hand, we distill only general abilities from teacher LLMs into the student SLMs of parameterized neural networks. On the other hand, for the specialized abilities and uncommon knowledge of a complex reasoning task, we employ a symbolic knowledge distillation approach to obtain and store the specialized knowledge within a symbolic knowledge base (KB).
By decoupling general and specialized capabilities, the proposed NesyCD can achieve superior performance cost-effectively, utilizing smaller models and blending parameterized neural networks with symbolic KB. Moreover, the specialized KB generalizes well and is comprehended and manipulated by humans.
Our experiments show that NesyCD significantly boosts SLMs' complex reasoning performance on in-domain (BBH, GSM8K) and out-of-domain (AGIEval, ARC) datasets. Notably, our approach enabled the LLaMA3-8B and Qwen2-7B to surpass GPT-3.5-turbo in performance and come close to matching LLaMA3-70B, despite the latter having $9\times$ more parameters. 



\end{abstract}

%
\begin{links}
    \link{Code}{https://github.com/Xnhyacinth/NesyCD}
    \link{Extended version}{https://arxiv.org/abs/2409.13203}
\end{links}

\section{Introduction}

Large Language Models (LLMs) \cite{Qwen2TR} excel in various complex reasoning tasks such as mathematical \cite{metamath}, commonsense \cite{csqa} and symbolic reasoning \cite{bbh} with In-Context Learning (ICL) \cite{iclcompositional} and Chain-of-Thought (CoT) Prompting~\cite{scaling}. Due to the high computational costs and expensive API calls required for LLMs (e.g., ChatGPT and LLaMA3-70B), enhancing Small Language Models (SLMs, e.g., $\leq$ 7B) to handle complex reasoning efficiently is more practical and crucial for large-scale deployment.


\begin{figure*}[t]
\centering
\includegraphics[width=2\columnwidth]{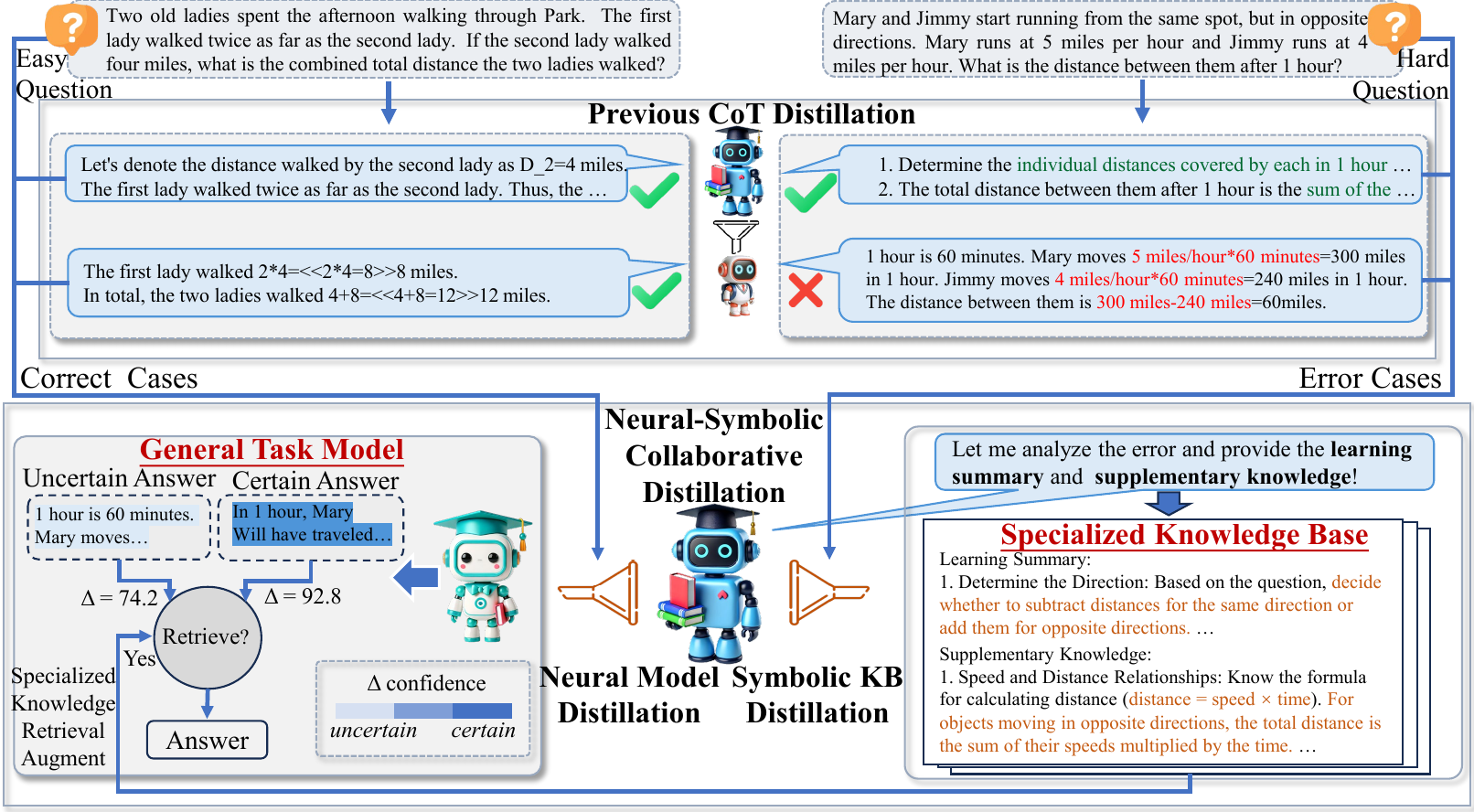} 
\caption{CoT distillation aims to train SLMs with the generated rationales obtained from LLMs, which is often limited by the SLMs' capabilities and frequently struggles to handle hard questions. The proposed NesyCD addresses this by decoupling the general and specialized knowledge of LLMs through SLMs' error analysis. It employs neural-based SLMs to model general knowledge while utilizing a symbolic specialized knowledge base (KB) to store specific knowledge. By adaptively utilizing the KB, NesyCD enhances the SLM's ability to handle complex reasoning tasks.} 
\label{intro}
\end{figure*}


To meet the practical needs mentioned above, many research efforts \cite{stdcot, distilling, mt-cot} in recent years have proposed the transfer of reasoning capabilities from teacher LLMs to student SLMs through CoT distillation (shown in the middle of Figure \ref{intro}). Specifically, LLMs generate high-quality rationales, which are then utilized to fine-tune SLMs.
This CoT distillation enhances the performance of SLMs in many tasks that require complex reasoning abilities such as arithmetic \cite{gsm8k} and symbolic reasoning \cite{chain}.

Despite some progress in CoT distillation, significant challenges persist that limit the performance of SLMs in complex reasoning tasks: 1) \textbf{Inconsistency in Capabilities between Teacher and Student Models}:
Existing methods often ignore the gap between the knowledge modeling and complex reasoning capabilities of LLMs and SLMs. Due to fewer parameters, student models struggle to acquire the comprehensive knowledge necessary for complex reasoning tasks \cite{kard}. As illustrated in Figure \ref{intro}, an SLM trained with traditional CoT distillation fails to solve hard questions that the teacher model can handle effectively.
2) \textbf{Difficulty in Modeling Sparse Specialized Knowledge}:
Neural knowledge distillation faces challenges in representing sparse and specialized knowledge due to the limited parameter space of distilled SLMs. In complex reasoning tasks, particularly with hard questions, specialized knowledge may contradict general knowledge (e.g., velocity superposition in relative motion). This sparse and specialized knowledge is hard to model in small-scale models, significantly impacting the SLM's performance on challenging questions and deteriorating its ability to handle unseen tasks.

In this paper, we argue that complex reasoning tasks require both general knowledge (e.g., numerical addition) and specialized knowledge (e.g., relative displacement). It can be simply understood as follows: the former refers to what SLMs can effectively model and is primarily used for answering high-frequency, easy questions, while the latter involves aspects that SLMs find challenging to model and are essential for addressing low-frequency, hard questions. For instance, as shown in Figure \ref{intro}, answering the right hard question demands domain-specific knowledge and advanced reasoning skills, like applying physics formulas to calculate displacement and understanding relative displacement.
To tackle the aforementioned challenges in existing CoT distillation methods, we propose a novel \textbf{Ne}ural-\textbf{Sy}mbolic \textbf{C}ollaborative \textbf{D}istillation (NesyCD) which transfers the general reasoning capabilities and common knowledge of LLMs to SLMs. Unlike using only neural-based models to build students before, we use a symbolic knowledge base (KB) to model and store relatively sparse specialized knowledge. Firstly, we gather correct and error cases made by SLMs fine-tuned with CoT distillation. Next, LLMs analyze error cases, extracting specialized knowledge through elaborate prompts and storing them in a symbolic KB. Finally, we fine-tune the SLMs with specialized knowledge augmented distillation, enhancing SLMs' abilities for hard questions. Moreover, to enhance the student SLMs' robustness against potentially noisy retrieved knowledge, we incorporate novel auxiliary tasks like augmented distillation (AD), answer prediction (AP) and direct CoT (DC) for multi-task learning to effectively utilize specialized knowledge.


To validate the effectiveness of NesyCD, we empirically demonstrate that it significantly improves the baseline performance of several open-source SLMs, such as TinyLLaMA \cite{TinyLlamaAO} and LLaMA2-7B \cite{llama}, across various benchmarks including GSM8k \cite{gsm8k} for mathematical reasoning, BBH \cite{bbh} and AGIEval \cite{AGIEval} for general reasoning, and ARC \cite{arc} for factual knowledge. Additionally, our extensive analysis shows that NesyCD is efficient regarding training data and model size. Specifically, the NesyCD-enhanced 1.1B TinyLLaMA outperforms the fine-tuned LLaMA2-7B and achieves superior results using only a quarter of the full training data compared to other strong baselines. 
Our findings and contributions are as follows:

\begin{itemize}
    \item We propose a neural-symbolic collaborative distillation method, which, to our knowledge, is the first approach to leverage a co-distillation framework that integrates neural-based models with symbolic knowledge bases for learning the complex reasoning capabilities of LLMs.
    \item We distinguish complex reasoning into general and specialized abilities through SLMs' error analysis. General abilities are modeled by a neural network, while specialized abilities are captured by a symbolic KB. Integrating these components enhances SLMs' complex reasoning, leading to more efficient models and reduced costs.
    \item The experimental results demonstrate that the proposed NesyCD significantly enhances the performance of SLMs across wide benchmarks for knowledge, mathematical, symbolic and other complex reasoning tasks both in-domain and out-of-domain.
\end{itemize}

\section{Related Work}

\subsection{CoT Distillation from LLMs}

The Chain-of-Thought (CoT) reasoning ability of LLMs, characterized by step-by-step question solving, is known as an emergent ability to improve performance in various reasoning tasks. Recent works \cite{LargeLM, Fu2023SpecializingSL} endeavor to transfer the CoT reasoning capabilities of LLMs to SLMs. Std-CoT \cite{stdcot} involves fine-tuning SLMs directly using CoTs extracted from teacher LLMs. Subsequent studies \cite{distilling, mt-cot} have proposed treating the learning of rationales and answers as separate optimization objectives. CasCoD \cite{improve} takes a different approach by decomposing the traditional single-step learning process into two cascaded steps.
However, the performance of these methods is hindered by the limited knowledge and capabilities of SLMs with fewer parameters \cite{kard, liao2025skintern}. This deficiency is particularly detrimental in complex reasoning tasks that require specialized knowledge and sophisticated reasoning skills. To address this issue, we propose to enhance SLMs by integrating knowledge retrieved from the specialized knowledge base (KB) generated by teacher LLMs.

\subsection{Knowledge-Augmented LMs}

Knowledge-augmented LMs (KALMs) enhance their reasoning by utilizing external KBs. A common approach involves retrieving relevant passages from sources like Wikipedia based on questions \cite{reading}. KARD \cite{kard} applies KALMs to knowledge-intensive tasks and finds it crucial for accurate answers and factual rationales. However, challenges like chunk indexing and independent encoding of documents can hinder KALMs' effectiveness in using external KB. To address this, we propose harnessing the world knowledge and reasoning capabilities of LLMs to generate specialized knowledge for KALMs, including learning summaries and supplementary knowledge to boost complex reasoning abilities.

\subsection{Learning from Errors}

Humans learn from their errors to avoid repeating them, and this capability has inspired efforts to enhance LLMs \cite{TurningDI, LearningFF}. LLM2LLM \cite{llm2llm} employs an instructor model to help target models learn from their errors. TRAN \cite{TongTRAN} uses a rule-based system to prevent past errors, while LEAP \cite{incontext} extracts and integrates principles from LLMs' errors into prompts. However, these approaches have not been adapted to improve SLMs, and the principles used in reasoning remain static rather than dynamically tailored based on the model's capabilities.

\section{Methods}

\begin{figure}[t]
\centering
\includegraphics[width=1.0\columnwidth]{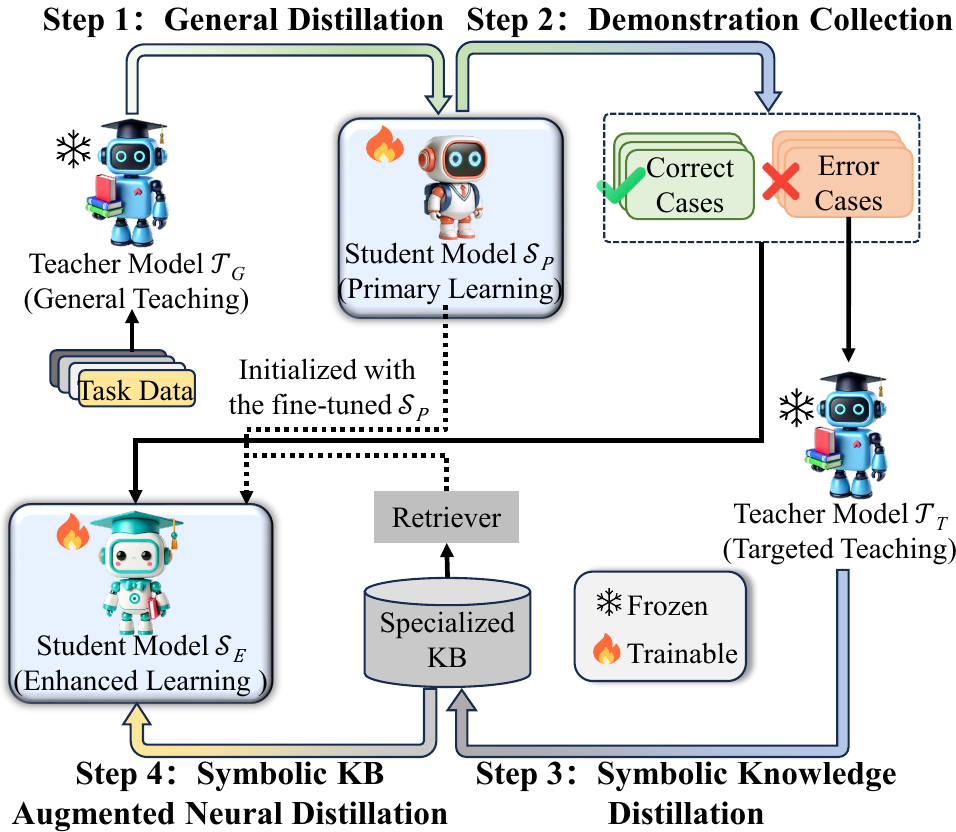} 
\caption{Overview of NesyCD.
1) General Distillation (§\ref{general_dis}): Fine-tune the student $\mathcal{S}_{P}$ to generate rationales obtained from the teacher $\mathcal{T}_{G}$ and answers.
2) Demonstration Collection (§\ref{demos_col}): Evaluate $\mathcal{S}_{P}$ and collect correct and error cases addressed by $\mathcal{S}_{P}$.
3) Symbolic Knowledge Distillation (§\ref{kb}): The teacher $\mathcal{T}_{T}$ analyzes errors and generate specialized KB.
4) Symbolic KB Augmented Neural Distillation (§\ref{nesy}): Use multi-task learning to fine-tune $\mathcal{S}_{E}$, enabling it to effectively utilize retrieved specialized knowledge.}
\label{model}
\end{figure}

We propose Neural-Symbolic Collaborative Distillation (NesyCD), which consists of four learning processes (illustrated in Figure \ref{model}):
1) \textbf{General Distillation} (§\ref{general_dis}), where a large language model (LLM) serves as a teacher model (General Teaching) $\mathcal{T}_{G}$ to generate rationales. Subsequently, a small language model (SLM) is fine-tuned to generate these rationales and provide answers to given questions, resulting in the Student Model (Primary Learning) $\mathcal{S}_{P}$;
2) \textbf{Demonstration Collection} (§\ref{demos_col}), where we evaluate the performance of the $\mathcal{S}_{P}$ on the specific task and dataset, and collect the correct and error cases for the following steps;
3) \textbf{Symbolic Knowledge Distillation} (§\ref{kb}), where the previous teacher model, acting as the Targeted Teaching model $\mathcal{T}_{T}$, analyzes and generates specialized knowledge aimed at the errors made by $\mathcal{S}_{P}$. This knowledge assists $\mathcal{S}_{P}$ in addressing similar tasks in the future and is then stored in a specialized knowledge base (KB).
4) \textbf{Symbolic KB Augmented Neural Distillation} (§\ref{nesy}), where the student model (Enhanced Learning) $\mathcal{S}_{E}$ is initialized with the fine-tuned $\mathcal{S}_{P}$. Using multi-task learning, fine-tune $\mathcal{S}_{E}$ to generate rationales and answers based on both questions and retrieved specialized knowledge, as well as on the questions alone.

\subsection{General Distillation}
\label{general_dis}

\noindent \textbf{Rationale Generation with LLMs:} The ability to generate high-quality rationales is known as the emergent ability of LLMs \cite{LargeLM}. Our objective is to transfer this capability to SLMs through CoT distillation. First, we employ CoT prompts \cite{chain} to guide the $\mathcal{T}_{G}$ in generating CoT.
We generate rationales for each training data point $\mathcal{D}_{\text{train}} = \left\{ (\boldsymbol{q}_i, \boldsymbol{a}_i) \right\}_{i=1}^n$, where $\boldsymbol{q}_i$ is a question and $\boldsymbol{a}_i$ is an answer, retaining only those that align with the correct answers in the dataset \cite{improve}.
\begin{equation}
    \boldsymbol{r}_{ij} = \mathcal{T}_{G}(\boldsymbol{p}, \boldsymbol{q}_i, \boldsymbol{a}_i)
\end{equation}
where $\boldsymbol{r}$ are generated rationales, $j \in \{1,...,l\}$ and $\boldsymbol{p}$ is the CoT prompt which is shown in Appendix \ref{prompt_cot}.

\noindent \textbf{Fine-tuning SLMs with Rationales:} We initially conduct Std-CoT \cite{stdcot} to develop the $\mathcal{S}_{P}$. Given a question $\boldsymbol{q}_i$, we fine-tune the $\mathcal{S}_{P}$ with trainable parameters $\theta$, to generate the rationale $\boldsymbol{r}_{ij}$ derived from the $\mathcal{T}_{G}$ and answer $\boldsymbol{a}_i$. We aim to minimize the negative log-likelihood of the sequence comprising the rationale $\boldsymbol{r}_{ij}$ and the answer $\boldsymbol{a}_i$, ensuring that the rationale precedes the answer.
\begin{equation}
    \mathcal{L}_{\text{Std-CoT}}(\theta) = - \frac{1}{n \cdot l} \sum_{i=1}^{n} \sum_{j=1}^{l} \log p_{\theta}(\boldsymbol{r}_{ij}, \boldsymbol{a}_i \mid \boldsymbol{q}_i)
\end{equation}

Rationales offer a deeper understanding of the reasoning behind answers, helping SLMs to respond more accurately \cite{distilling}. However, SLMs with limited parameters may struggle to retain all training data and complex reasoning capabilities, which can affect the quality of rationale generation \cite{kard}. Furthermore, this implicit learning may cause SLMs to focus on answering questions directly after reading, potentially impairing generalization in reasoning \cite{improve}. Therefore, it is essential to assess the knowledge and capabilities that SLMs fail to acquire and have teacher LLMs generate specialized knowledge. This approach enables SLMs to retrieve and utilize knowledge effectively, enhancing their ability to produce high-quality rationales and perform complex reasoning when needed.

\subsection{Demonstration Collection}
\label{demos_col}
The \(\mathcal{S}_{P}\) analyzes \(\mathcal{D}_{\text{train}}\) to generate predicted rationales \(\hat{\boldsymbol{r}}\) and answers \(\hat{\boldsymbol{a}}\). Incorrect solutions are identified by comparing each \(\hat{\boldsymbol{a}_i}\) with the actual answer \(\boldsymbol{a}_i\). The collected errors, \(\mathcal{D}_{\text{neg}}\), reveal the weaknesses of the student model.

\begin{equation}
    \mathcal{D}_{\text{neg}} = \left\{ (\boldsymbol{q}_i, \hat{\boldsymbol{r}}_i, \boldsymbol{a}_i) \mid \hat{\boldsymbol{a}}_i \neq \boldsymbol{a}_i, (\boldsymbol{q}_i, \boldsymbol{a}_i) \in \mathcal{D}_{\text{train}} \right\}
\end{equation}

In our view, correct cases are those manageable by conventional distillation models using general knowledge. In contrast, error cases are difficult for small-scale neural models to handle effectively, requiring additional specialized knowledge and advanced reasoning capabilities. This specialized knowledge is sparse, making it more cost-effective to represent through a symbolic KB, as neural networks would need a larger parameter scale to capture it.

\subsection{Symbolic Knowledge Distillation}
\label{kb}

The $\mathcal{T}_{T}$ examines each error $\hat{\boldsymbol{r}}_i$ in $\mathcal{D}_{\text{neg}}$ and generates specialized knowledge $\boldsymbol{k}_i$, including generalized learning summaries $\boldsymbol{k}_i^m$ and supplemental knowledge $\boldsymbol{k}_i^p$. This addresses the issues of insufficient knowledge and lack of reasoning ability in the SLM. For each question $\boldsymbol{q}_i$ in $\mathcal{D}_{\text{neg}}$, we construct the teacher model’s prompt\footnote{The prompt for generating $\boldsymbol{k}_i$ is in the Appendix \ref{prompt_know}.} $\boldsymbol{p'}$ that incorporates the student's incorrect rationale $\hat{\boldsymbol{r}_i}$, and the correct answer $\boldsymbol{a}_i$ which constructs a specialized KB, represented as $\mathcal{D}_{\text{k}} = \{(\boldsymbol{q}_i, \boldsymbol{k}_i) ~|~ (\boldsymbol{q}_i, \hat{\boldsymbol{r}_i}, \boldsymbol{a}_i) \in \mathcal{D}_{\text{neg}}$. The process is primarily driven by the identification and correction of errors.
\begin{equation}
    \boldsymbol{k}_{i} = \mathcal{T}_{T}(\boldsymbol{p'}, \boldsymbol{q}_i, \hat{\boldsymbol{r}}_i, \boldsymbol{a}_i)
\end{equation}
\subsection{Symbolic KB Augmented Neural Distillation}
\label{nesy}

Inspired by knowledge augmentation \cite{kard}, we propose retrieving relevant specialized knowledge from the specialized KB which is generated through error analysis of the SLM to support its memory and reasoning capabilities. Acquiring specialized knowledge is crucial for training the SLM to produce high-quality rationales, subsequently leading to the correct answers to given questions. In alignment with prior knowledge-intensive tasks, we employ a dense retriever Contriever \cite{contriever} to retrieve a set of relevant questions for each question: $\mathcal{Q}_{i} = \text{topk}(\rho(\boldsymbol{q}~|~\boldsymbol{q}_i; \mathcal{D}_{\text{k}}), m)$, where $\rho$ scores the questions $\boldsymbol{q} \in \mathcal{D}_{\text{k}}$ based on their relevance to the question $\boldsymbol{q}_i$, and $\text{topk}$ selects the top $m$ questions with the highest relevance scores. Then we can get the specialized knowledge:
\begin{equation}
    \mathcal{K}_{i} = \{\boldsymbol{k}_{j}~|~ (\boldsymbol{q}_j, \boldsymbol{k}_j) \in \mathcal{D}_{\text{k}}, \boldsymbol{q}_j \in \mathcal{Q}_{i} \}
\end{equation}
Finally, we fine-tune the student model $\mathcal{S}_{E}$, initialized with the fine-tuned student $\mathcal{S}_{P}$, using the retrieved specialized knowledge $\mathcal{K}_{i}$ to generate the rationale $\boldsymbol{r}_{ij}$ and the answer $\boldsymbol{a}_i$ for the question $\boldsymbol{q}_i$.
\begin{equation}
    \mathcal{L}_{\text{NesyCD}}(\theta) = - \frac{1}{n \cdot l} \sum_{i=1}^{n} \sum_{j=1}^{l} \log p_{\theta}(\boldsymbol{r}_{ij}, \boldsymbol{a}_i \mid \boldsymbol{q}_i, \mathcal{K}_{i})
\end{equation}
where the rationale and answer are sequentially generated as we did in §\ref{general_dis}. Beyond fine-tuning the SLM with \textbf{augmented distillation} (AD), we propose two auxiliary tasks in \textbf{multi-task learning} to enhance reasoning capabilities. These tasks aim to improve the SLM's ability to integrate and apply specialized knowledge effectively: 1) \textbf{Answer Prediction} (AP), which generates answers directly, aiming to help SLMs internalize the reasoning required for direct questions that do not necessitate a CoT; 2) \textbf{Direct CoT} (DC), which relies solely on the SLM’s intrinsic knowledge (i.e., $\mathcal{K}_i$ is empty) to address relatively easy questions.



During the inference stage, as illustrated at the bottom of Figure \ref{intro}, we determine the necessity of retrieval based on the model confidence \cite{ChainofThoughtRW}. For instances requiring retrieval, we extract the most specialized knowledge from the specialized knowledge base relevant to the question to assist in generating the rationale and answer.
\begin{equation}
    \Delta_{\text{answer}} = \frac{1}{|\text{answer}|} \sum_{x_t \in \text{answer}} p(x_{t}^1 \mid x_{<t}) - p(x_{t}^2 \mid x_{<t})
\end{equation}
Here, $x_{t}^1$ and $x_{t}^2$ represent the top two tokens at the $t$-th decoding step. These tokens are selected based on their highest post-softmax probabilities from the vocabulary, given that $x_t$ is part of the answer tokens. More analysis about $\Delta_{\text{answer}}$ can be seen in Appendix \ref{confi}.

\begin{table*}[t]
\small
  \centering
  \setlength{\tabcolsep}{1.5mm}
      \begin{tabular}{lcccccccc}
        \toprule
         \multirow{2}{*}{\textbf{Methods}} & \multicolumn{2}{c}{\textbf{In-Domain}}& \multicolumn{5}{c}{\textbf{Out-Of-Domain}} & \multirow{2}{*}{\textbf{Average}}\\
        \cmidrule(r){2-3}\cmidrule(r){4-8}
         & \textbf{BBH-test} & \textbf{GSM8K} & \textbf{BB-sub} & \textbf{AGIEval} & \textbf{GSM8K-PLUS} & \textbf{ARC-E}  & \textbf{ARC-C} &  \\
        \midrule
        \multicolumn{9}{l}{\textit{\# Closed-source model and Open-source models (Zero-shot-CoT)}}\\
        GPT-3.5-turbo (\textit{Teacher}) & 43.2 & 72.6 & 44.0 & 50.5 & 55.9 & 91.8 & 84.1 & 63.2 \\
        LLaMA-3-70B-Instruct & 62.6 & 89.2 & 51.0 & 66.3 & 72.9 & 97.6 & 93.2 & 76.1\\
        \midrule
        \multicolumn{9}{l}{\textit{\# TinyLLaMA-1.1B based}}\\
        Zero-shot \cite{zeroshot}        & 14.0 & 2.0 & 17.7 & 17.8 & 1.5 & 19.4 & 15.0 & 12.5\\
        Zero-shot-CoT \cite{zeroshotcot} & 13.5 & 1.4 & 17.7 & 10.4 & 1.3 & 16.0 & 13.4 & 10.5 \\
        \hdashline
        Fine-tuning                     & 48.8 & 3.5 & 26.0 & 21.2 & 3.7 & 28.0 & 24.6 & 22.3\\
        Knowledge-Augmented Fine-tuning & 49.3 & 3.7 & 27.4 & 21.9 & 3.3 & 29.4 & 25.3 & 22.9\\
        \hdashline
        Std-CoT \cite{stdcot} & 47.8$_{\pm .43}$ & ~~7.9$_{\pm .27}$ & 27.6$_{\pm .31}$ & 21.5$_{\pm .56}$ & 4.3$_{\pm .62}$ & 28.2$_{\pm .69}$ & 25.0$_{\pm .48}$ & 23.2 \\
        MT-CoT \cite{mt-cot} & 44.1$_{\pm .78}$ & ~~4.1$_{\pm .35}$ & 25.0$_{\pm .45}$ & 21.4$_{\pm .64}$ & 2.8$_{\pm .83}$ & 33.5$_{\pm .52}$ & 25.1$_{\pm .59}$ & 22.3 \\
        Step-by-step \cite{distilling} & 42.4$_{\pm .56}$ & ~~4.3$_{\pm .47}$ & 26.2$_{\pm .38}$ & 21.1$_{\pm .72}$ & 3.1$_{\pm .54}$ & 29.6$_{\pm .61}$ & 25.9$_{\pm .66}$ & 21.8 \\
        KARD (BM25) \cite{kard} & 49.5$_{\pm .61}$ & ~~7.6$_{\pm .40}$ & 26.9$_{\pm .43}$ & 20.2$_{\pm .48}$ & 4.0$_{\pm .77}$ & 28.2$_{\pm .85}$ & 26.5$_{\pm .91}$ & 23.3 \\
        CasCoD \cite{improve} & 48.1$_{\pm .49}$ & ~~6.8$_{\pm .39}$ & 23.1$_{\pm .64}$ & 19.4$_{\pm .73}$ & 4.8$_{\pm .48}$ & 29.0$_{\pm .63}$ & 27.1$_{\pm .42}$ & 22.6 \\
        \textbf{NesyCD} (\textit{ours}) & \textbf{66.3$_{\pm .42}$} & \textbf{11.8$_{\pm .83}$} & \textbf{30.6$_{\pm .27}$} & \textbf{23.1$_{\pm .41}$} & \textbf{7.2$_{\pm .93}$} & \textbf{36.2$_{\pm .76}$} & \textbf{29.0$_{\pm .58}$} & \textbf{29.3} \\
        \midrule
        \multicolumn{9}{l}{\textit{\# LLaMA2-7B based}}\\
        Zero-shot \cite{zeroshot}        & 17.3 & 2.7 & 18.6 & 19.2 & 2.4 & 25.2 & 20.6 & 17.0  \\
        Zero-shot-CoT \cite{zeroshotcot} & 13.5 & 3.1 & 12.2 & 10.3 & 2.1 & 29.1 & 20.2 & 12.9 \\
        \hdashline
        Fine-tuning                     & 57.8 & 5.8 & 33.3 & 31.0 & 5.8 & 73.3 & 56.3 & 37.6\\
        Knowledge-Augmented Fine-tuning & 58.7 & 6.3 & 34.2 & 31.8 & 6.1 & 75.1 & 57.0 & 38.5\\
        \hdashline
        Std-CoT \cite{stdcot}  & 58.1$_{\pm.74}$ & 20.5$_{\pm.71}$  & 30.7$_{\pm.48}$ & 23.6$_{\pm.65}$    & 12.0$_{\pm.26}$   & 73.4$_{\pm.81}$  & 55.9$_{\pm.78}$  & 39.2   \\
        MT-CoT \cite{mt-cot} & 46.4$_{\pm .52}$ & ~~7.5$_{\pm .48}$ & 28.1$_{\pm .55}$ & 32.1$_{\pm .53}$ & ~~5.8$_{\pm .39}$ & 70.3$_{\pm .67}$ & 55.7$_{\pm .45}$ & 35.1 \\
        Step-by-step \cite{distilling} & 53.9$_{\pm .69}$ & ~~8.3$_{\pm .57}$ & 32.3$_{\pm .33}$ & 32.4$_{\pm .40}$ & ~~5.6$_{\pm .41}$ & 74.9$_{\pm .52}$ & 60.0$_{\pm .56}$ & 38.2 \\
        KARD (BM25) \cite{kard} & 59.2$_{\pm .93}$ & 23.5$_{\pm .62}$ & 30.8$_{\pm .66}$ & 29.2$_{\pm .79}$ & 15.2$_{\pm .54}$ & 70.2$_{\pm .71}$ & 55.4$_{\pm .48}$ & 40.5 \\
        CasCoD \cite{improve} & 59.6$_{\pm .78}$ & 23.6$_{\pm .87}$ & 32.2$_{\pm .71}$ & 28.8$_{\pm .63}$ & 14.5$_{\pm .68}$ & 72.6$_{\pm .49}$ & 56.7$_{\pm .83}$ & 41.1 \\
        \textbf{NesyCD} (\textit{ours}) & \textbf{75.5}$_{\pm.69}$ & \textbf{32.4}$_{\pm.53}$ & \textbf{36.9}$_{\pm.38}$ & \textbf{33.6}$_{\pm.71}$ & \textbf{24.1}$_{\pm.47}$ & \textbf{77.5}$_{\pm.89}$ & \textbf{60.8}$_{\pm.56}$ & \textbf{48.7} \\
        \bottomrule
      \end{tabular}
    \caption{Performance (\%) of LLaMA2-7B \cite{llama} and TinyLLaMA-1.1B \cite{TinyLlamaAO} with different methods across seven selected datasets. \textbf{Bold} indicates the best in each setting. We report the mean and standard deviation of accuracy with 3 different runs for CoT distillation methods. We provide a systematic case study in Appendix \ref{case}.}
    \label{main}
\end{table*}

\section{Experiments}

In this section, we conduct extensive experiments and comprehensive analysis to evaluate the effectiveness of NesyCD on both in-domain (ID) and out-of-domain (OOD) datasets.

\begin{figure*}[t]
\centering
\includegraphics[width=1.7\columnwidth]{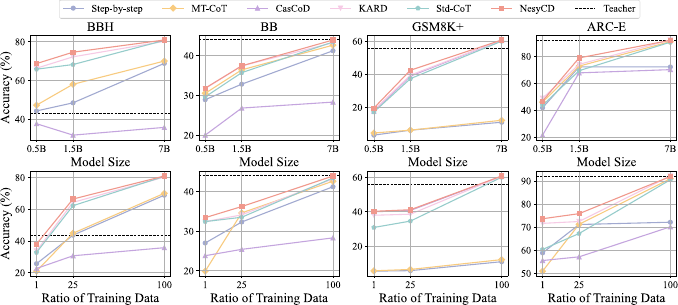} 
\caption{Efficiency on training data and model size. The backbone model for the data size variation is Qwen2-1.5B.}
\label{size}
\end{figure*}

\subsection{Datasets}

Following \cite{wang2023how, llms}, we focus on three practical abilities: factual, mathematical, and general reasoning. For each ability, we select a relevant public dataset, integrate its training data into the target dataset \(\mathcal{D}_{\text{train}}\) for mixed training, and combine its test data into the evaluation dataset \(\mathcal{D}_{\text{eval}}\). Additionally, each ability includes an OOD dataset in \(\mathcal{D}_{\text{eval}}\). This setup allows us to evaluate the model's ability to generalize and enhance performance beyond the ID training environment.

\noindent \textbf{Factual Reasoning}: We select the Multitask Language Understanding (MMLU) \cite{mmlu} as the ID dataset, which includes multiple-choice questions across 57 subjects. For OOD evaluation, we use the ARC \cite{arc}, comprising both Easy and Challenge segments.

\noindent \textbf{Mathematical Reasoning}: We select MetaMathQA \cite{metamath} as the ID dataset, which includes a high-quality collection of forward and reverse mathematical reasoning question-answer pairs, derived from GSM8K \cite{gsm8k} and MATH \cite{math}. For OOD evaluation, we use GSM8K and GSM8K+ \cite{gsmplus}.

\noindent \textbf{General Complex Reasoning}: We chose BIG-Bench Hard (BBH) \cite{bbh} as the ID dataset, which includes 27 challenging tasks spanning arithmetic, symbolic reasoning, and more, derived from BIG-Bench (BB) \cite{bb}. Most of the data consists of multiple-choice questions. For OOD evaluation, we use BB-Sub filtered by CasCoD, and AGIEval \cite{AGIEval} subtasks about English multiple-choice questions.


\subsection{Baselines}

We compare our method with the following baselines: \textit{1)} \textbf{Teacher \& Vanilla Student} in Zero-shot \cite{zeroshot}, Zero-shot-CoT \cite{zeroshotcot}. \textit{2)} \textbf{Fine-tuning} involves fine-tuning a model to generate answers given only questions. The performance of the baselines above illustrates the capability of SLMs to solve tasks using only training data, without external guidance or additional knowledge. \textit{3)} \textbf{CoT distillation} includes \textbf{Std-CoT} \cite{stdcot} which is the standard CoT distillation method, enabling direct fine-tuning of the student model with CoT data; \textbf{Step-by-step} \cite{distilling} is a multi-task method that extracts rationales and answers separately; \textbf{MT-CoT} \cite{mt-cot} is another multi-task method that optimizes both answer prediction and CoT generation simultaneously; \textbf{CasCoD} \cite{improve} decomposes the traditional single-step learning process into two cascaded learning steps. \textit{4)} \textbf{Knowledge-Augmentation} involves attaching retrieved passages to the question during both training and inference. This includes \textbf{Knowledge-Augmented Fine-tuning} focuses on generating answers only, and \textbf{KARD} \cite{kard} emphasizes learning the generation of rationales.

\subsection{Implementations}

For all experiments, we use the LLaMA2-7B \cite{llama} and TinyLLaMA-1.1B \cite{TinyLlamaAO} as the student SLM. We query the teacher model GPT-3.5-turbo to annotate the CoTs data with the manual prompt \cite{bbh}. Unless otherwise specified, $m$ is set to 1 (§\ref{number}) and $\Delta_\text{threshold}$ is set to 0.68 (§\ref{impact}). We follow the standard metrics and datasets statics elaborated in Appendix \ref{app_data}.

We employ LoRA \cite{hu2022lora} for parameter-efficient fine-tuning of the student SLMs. All experiments are conducted on 2 A100 GPUs with 80GB. During the inference stage, we utilize vLLM \cite{vllm} to accelerate inference. Detailed information about training and hyperparameters is provided in Appendix \ref{hyperp}.

\begin{table}[t]
\centering
\small
\setlength{\tabcolsep}{1.2mm}
\begin{tabular}{lccccc}
\hline
\textbf{Retriever} & \textbf{BBH} & \textbf{BB} & \textbf{AGIEval} & \textbf{GSM8K+} & \textbf{ARC-E} \\
\hline
Contriever & 75.49  & 36.92 & \textbf{33.63} & \textbf{24.11} & \textbf{77.53} \\
DPR & 74.47  & 36.84 & 32.87 & 23.43 & 77.48 \\
\hdashline
BM25 & \textbf{75.55}  & \textbf{37.11} & 31.44 & 23.57 & 77.10 \\
\hline
\end{tabular}
\caption{Results for different retrievers.}
\label{table:retrievers}
\end{table}

\subsection{Main Results}

Table \ref{main} shows that the NesyCD has \textbf{achieved significant improvements on both ID and OOD datasets} using two weaker SLMs\footnote{More results about LLaMA3 and Qwen2 are in Appendix \ref{exre}.}. Specifically, LLaMA2-7B and TinyLLaMA-1.1B demonstrated an average improvement of 8.4\% and 5.9\% respectively, consistently outperforming all existing baselines. For an analysis of model size, please refer to §\ref{model_size}. The impact of NesyCD decreases as the model size (and hence capability) increases because larger models can retain knowledge better during pre-training and fine-tuning.

Compared to Zeroshot and Zeroshot-CoT, \textbf{CoT distillation has significantly improved the performance of SLM}. While fine-tuning methods can significantly enhance factual and general reasoning abilities, \textbf{fine-tuning's impact on mathematical reasoning remains minimal}. CoT distillation aids SLM in generating rationales that clarify intermediate steps, thereby enhancing mathematical reasoning capabilities. Among various CoT distillation methods, NesyCD not only boosts symbolic knowledge learning to rectify SLM errors but also adaptively retrieves specialized knowledge based on confidence during tests to assist in producing high-quality rationales, thus improving overall performance. In comparison to KARD \cite{kard}, \textbf{our specialized KB is more relevant to the capabilities of the SLM and the question at hand}. Additionally, it generates a distribution more aligned with the SLM through the LLM, significantly enhancing effectiveness \cite{generate}. Our NesyCD significantly enhances the SLM's performance, demonstrating the effectiveness of neural-symbolic knowledge integration in complex reasoning tasks.




\begin{table}[t]
\small
\setlength{\tabcolsep}{1.2mm}
\centering
\begin{tabular}{lccccc}
\hline
\textbf{\# Knowledge} & \textbf{BBH} & \textbf{BB} & \textbf{AGIEval} & \textbf{GSM8K+} & \textbf{ARC-E} \\
\hline
$m = 1$ & \textbf{75.49}  & \textbf{36.92} & \textbf{33.63} & \textbf{24.11} & \textbf{77.53} \\
$m = 2$ & 72.41  & 35.36 & 32.57 & 21.28 & 77.48 \\
$m = 3$ & 73.28  & 34.71 & 31.83 & 20.92 & 76.98 \\
\hline
\end{tabular}
\caption{Results for different $m$.}
\label{table:number}
\end{table}

\subsection{Efficiency on Dataset and Model Sizes}
\label{model_size}

To evaluate the efficiency of NesyCD in terms of training data and model size, we measured test accuracy using Qwen2's \cite{Qwen2TR} 0.5B, 1.5B, and 7B models across various methods while varying the amount of training data and model size. As shown at the bottom of Figure \ref{size}, NesyCD successfully transfers the reasoning capabilities of the teacher LLM by generating symbolic specialized knowledge, even with minimal training data. As the training data decreases, the performance gap between NesyCD and other baselines widens, demonstrating NesyCD's superior robustness and sample efficiency. \textbf{This suggests that NesyCD performs better with fewer samples and that its effectiveness can be further enhanced by increasing the training data, allowing for more effective distillation}.

Regarding model size efficiency, as shown at the top of Figure \ref{size}, NesyCD outperforms other baselines across various model scales. Notably, NesyCD enables Qwen2-7B to surpass the teacher GPT-3.5 Turbo in both ID and OOD performance, despite having over $10\times$ fewer parameters. 
These results highlight NesyCD's significant practical benefits in resource-constrained environments, as it reduces the computational cost required for SLMs while achieving performance levels that exceed those of larger LLMs. \textbf{This further illustrates that SLMs cannot fully utilize the CoT reasoning generated by LLMs, thereby necessitating the implementation of our proposed NesyCD}.

\begin{figure}[t]
\centering
\includegraphics[width=0.95\columnwidth]{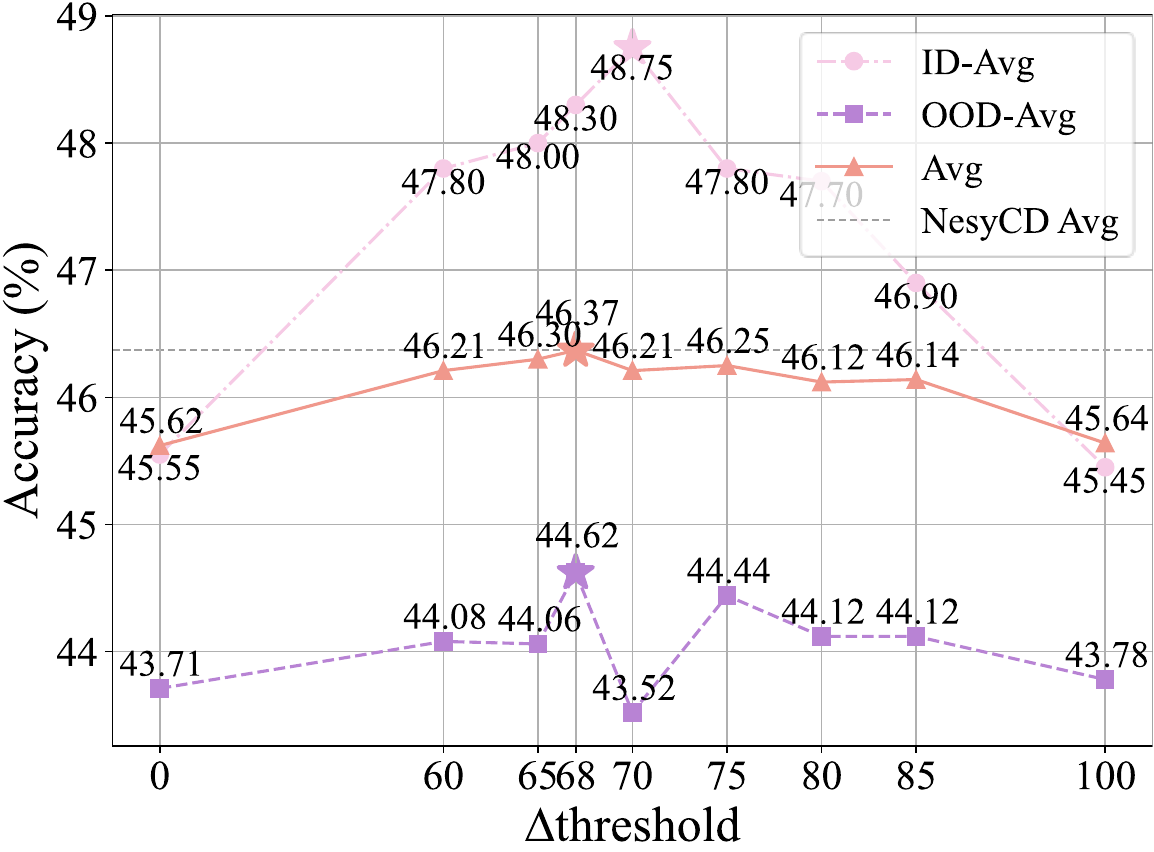}
\caption{Performance variation trend on \(\Delta_{\text{threshold}}\). The results are reported by ID-Avg and OOD-Avg which respectively denote average accuracy on ID and OOD datasets.}
\label{thre}
\end{figure}

\subsection{Performances with Different Retrievers}
\label{retrieval}

We examined the impact of various retrievers including DPR \cite{DensePR}, Contriever \cite{contriever}, and BM25 \cite{bm25} on the performance of NesyCD using LLaMA2-7B. As shown in Table \ref{table:retrievers}, the performance differences among these retrievers are minimal, with Contriever performing slightly better. This finding suggests that NesyCD can attain improved benefits as the quality of the retriever advances, effectively retrieving more relevant specialized knowledge and enhancing the capability to solve complex reasoning tasks.

\subsection{The Number of Knowledge Used for Inference}
\label{number}

Even LLMs can be easily distracted by irrelevant background information \cite{Shi2023LargeLM} or extended context \cite{LostIT}. Therefore, simply adding more knowledge during the inference process does not necessarily enhance performance unless the relevant knowledge is selected. Table \ref{table:number} illustrates the impact of the number of knowledge used during inference in the NesyCD ($m$ in §\ref{number}) on LLaMA2-7B. We observe that performance decreases as $m$ increases which implies that including additional knowledge does not always enhance reasoning and can interfere with the model's judgment, corroborating previous research findings.

\subsection{Impact of Confidence Threshold}
\label{impact}
We investigated the effect of varying the confidence threshold \(\Delta_{\text{threshold}}\) on performance across both ID and OOD datasets using LLaMA2-7B, as shown in Figure \ref{thre}. A higher confidence threshold means the model requires greater certainty to trust its output. 
In extreme cases, \(\Delta_{\text{threshold}} = 0\) means no retrieval is performed, while \(\Delta_{\text{threshold}} = 100\) means retrieval is performed for all cases. Both extremes lead to significant performance degradation, underscoring the need for adaptive retrieval. For easy tasks, the model might produce hallucinations and incorrect answers with additional knowledge, while for complex tasks lacking external guidance, the model cannot rely solely on internal parameters. Despite these challenges, our method consistently outperforms other baselines, even in extreme scenarios. 


\subsection{Ablation Studies}

\begin{table}[t]
\small
\centering
\setlength{\tabcolsep}{1.2mm}
\begin{tabular}{lccccc}
\hline
\textbf{Methods} & \textbf{BBH} & \textbf{BB} & \textbf{AGIEval} & \textbf{GSM8K+} & \textbf{ARC-E} \\
\hline
\textbf{NesyCD} & \textbf{75.49}  & \textbf{36.92} & \textbf{33.63} & \textbf{24.11} & \textbf{77.53} \\
w/o $\boldsymbol{k}^m$ & 68.48 & 35.32 & 31.75 & 20.93 & 76.86 \\
w/o $\boldsymbol{k}^p$ & 68.86  & 35.64 & 31.59 & 22.35 & 77.06 \\
w/o AD & 64.34  & 32.23 & 28.85 & 20.43 & 74.79\\
w/o AP \& DC & 66.51  & 35.12 & 29.83 & 21.88 & 77.41\\
\hline
Std-CoT             & 58.13  & 30.68 & 23.61 & 12.02 & 73.36\\
w $\boldsymbol{k}$ & 61.24  & 31.33 & 26.41 & 13.75 & 74.32\\
\hline

Zero-shot-CoT & 13.51  & 12.19 & 10.32 & 2.08 & 29.13\\
w $\boldsymbol{k}$ & 14.57  & 13.23 & 11.68 & 2.92 & 30.68\\

\hline
\end{tabular}
\caption{Ablation studies on different components.}
\label{table:ab}
\end{table}

To demonstrate the effectiveness of NesyCD, we created four variants by individually removing the learning summary ${\boldsymbol{k}^m}$, the supplementary knowledge ${\boldsymbol{k}^p}$, the augmented distillation (AD), and the multi-task learning (AP \& DC, §\ref{nesy}) respectively. Specialized knowledge $\boldsymbol{k}$ is composed of ${\boldsymbol{k}^m}$ and ${\boldsymbol{k}^p}$ (§\ref{demos_col}). We employed LLaMA2-7B as the SLM for ablation studies, and the results are presented in Table \ref{table:ab}. We can observe that performance diminishes with the exclusion of any single component, underscoring the significance of each element. Additionally, specialized knowledge exhibits orthogonality and universality, enhancing Zero-shot-CoT and other CoT distillation methods (w $\boldsymbol{k}$) which confirms the importance of refining symbolic knowledge.

\section{Conclusion}
In this work, we introduce Neural-Symbolic Collaborative Distillation (NesyCD), a method aimed at enhancing the capabilities of Small Language Models (SLMs) for complex reasoning tasks that require additional knowledge and advanced reasoning skills. NesyCD uses Large Language Models (LLMs) to analyze SLM errors and generate specialized knowledge, including learning summaries and supplementary knowledge, organized into an external knowledge base. By integrating parameter updates with retrieving specialized knowledge, NesyCD improves both rationale generation and answer accuracy for SLMs. Empirical experiments show that NesyCD surpasses fine-tuning and CoT distillation baselines in in- and out-of-domain scenarios. 

\section*{Acknowledgements}

This work was supported by the National Key R\&D Program of China (No. 2022ZD0160503) and the National Natural Science Foundation of China (No.62376270, No.62276264).

\bibliography{aaai25}

\appendix

\section{Experimantal Settings}
\label{experiment}

\subsection{Datasets}
\label{app_data}

Table \ref{stat_data} shows the statistics details of the selected datasets.

\begin{table*}[htbp]
  \centering
  \renewcommand\arraystretch{1.05}
      \begin{tabular}{llccc}
        \toprule
        \textbf{Abilities} & \textbf{Task} & \textbf{\# Train} & \textbf{\# Train (Filtered)} & \textbf{\# Test} \\
        \midrule
        \multirow{3}{*}{Factuality} & ID: MMLU & Dev + Val: 1,815 & 1,556 & - \\
                                    & OOD: ARC-C & -  & - & 1,172 \\
                                    & OOD: ARC-E & - & - & 2,376 \\
                                    \hline
        \multirow{3}{*}{Math} & ID: MetaMathQA & 395,000  & 3,500 & - \\
                                    & ID: GSM8K & -  & - & 1,319 \\
                                    & OOD: GSM8K-PLUS & - & - & 1,400 \\
                                    \hline
        \multirow{3}{*}{Reasoning} & ID: BBH & 6,511  & 3,805 & 1,304 \\
                                    & OOD: BB-sub & -  & - & 5,384 \\
                                    & OOD: AGIEval & - & - & 2,546 \\
                                    \hline
        \textbf{All} & \textbf{Sum} & - & 8,860 & 15,501\\
        \bottomrule
      \end{tabular}
    \caption{Statistical details of the selected datasets. Since MMLU lacks official training data, we combined the development and validation datasets to form a training set. To maintain sample balance, we matched the size of MetaMathQA to that of BBH. We obtained balanced samples from two dataset augmentation modes, MATH\_Aug and GSM\_Aug, resulting in a total of 3,500 samples.}
    \label{stat_data}
\end{table*}

For MMLU \cite{mmlu}, we adhere to previous prompt styles \cite{bbh}, instructing the teacher model (e.g., GPT-3.5-Turbo) to generate answers and Chain of Thought (CoT). By excluding samples with incorrect answers, we ultimately obtained a total of 1,556 samples. For MetaMathQA \cite{metamath}, we acquired 3,500 samples through random sampling. For BB \cite{bb}, we followed the CasCoD \cite{improve} methodology by filtering the original dataset for tasks containing the keyword "multiple choice" and randomly extracting up to 100 examples for each task. Note that tasks in BBH do not involve BB-sub. 

During the evaluation stage, we use Exact Match \cite{squad} as the evaluation metric.

The answer generation between the involved models is conducted in a zero-shot setting, with all models set to a temperature of 0.8 and a maximum token length of 1024. The prompt can be found in the Appendix \ref{prompt_cot}.

\begin{table*}[htbp]

  \centering
  \renewcommand\arraystretch{1.05}
   \resizebox{\linewidth}{!}{
      \begin{tabular}{lccccccc}
        \toprule
        \textbf{Hyperparameter}  & \textbf{TinyLLaMA-1.1B} &  \textbf{LLaMA2-7B} & \textbf{LLaMA3-8B}& \textbf{Qwen2-0.5B}& \textbf{Qwen2-1.5B}& \textbf{Qwen2-7B}& \textbf{NesyCD}\\
        \midrule
        Max Input Len & 2048 & 4096 & 4096 & 4096 & 4096 & 4096 & - \\
        Max Output Len & 128 & 128 & 128 & 128 & 128 & 128  & -\\
        Optimizer & AdamW & AdamW  & AdamW & AdamW & AdamW & AdamW& AdamW \\
        Learning Rate & 2e-4 & 1e-4  & 5e-5 & 2e-4 & 1e-4 & 1e-4 & - \\
        precision & fp16 & fp16 & fp16 & fp16 & fp16 & fp16 & fp16\\
        \# Training epochs & 10 & 10 & 10  & 10 & 10 & 10 & 10\\
        \# Warmup Steps & \multicolumn{6}{c}{10\% of total training steps} & - \\
        Batch Size & 32 & 16 & 8 & 32 & 16 & 8 & 8  \\
        Gradient Accumulation & 1 & 2 & 4 & 1 & 2 & 4 & 4 \\
        rank of LoRA & 32 & 32 & 32 & 32 & 32 &32 & -\\
        \bottomrule
      \end{tabular}
      }
        \caption{Training hyperparameters.}
          \label{hyperparam}
\end{table*}

\begin{table}[htbp]

  \centering
  \renewcommand\arraystretch{1.05}
   \resizebox{\linewidth}{!}{
      \begin{tabular}{lccc}
        \toprule
        \multirow{2}{*}{\textbf{Hyperparameter}} & \multirow{2}{*}{\textbf{Student}} & \multicolumn{2}{c}{\textbf{Teacher}}\\
        \cmidrule(r){3-4}
        & & Rationale & Reasoning\\
        \midrule
        do\_sample & False & True & False\\
        temperature & 0.6 & 0.8 & 0.6 \\
        top-p & 0.95 & 1.0 & 0.95 \\
        top-k & 50 & 50 & 50\\
        max\_new\_tokens & 1024 & 2048 & 1024 \\
        \# return sequences & 1 & 2 & 1 \\
        \bottomrule
      \end{tabular}
      }
        \caption{Generation configs of students and teachers.}
          \label{gen}
\end{table}

\subsection{Hyperparameter}
\label{hyperp}
The complete set of stable hyperparameters used for both baseline models and the proposed NesyCD training and inference runs can be found in Table \ref{hyperparam} and Table \ref{gen}, respectively.

In our research, we ensured consistent hyperparameter settings across all baselines, including the proposed NesyCD method, to maintain the fairness of our comparative analysis. Detailed hyperparameters and their explanations are presented below. For NesyCD, particularly in the fourth step, we reduced the enhanced distillation parameters to 3 epochs and fixed the batch size at 8, as the concatenation of specialized knowledge results in longer inputs. We maintained a consistent batch size across all baselines to eliminate any performance differences attributable to varying batch sizes, which depend on model size, with larger models necessitating smaller batch sizes. The learning rate, a key parameter affecting model performance, was set to 5e-5, 1e-4, 2e-4, and 3e-4 in a series of experiments, revealing that larger models require smaller learning rates. Consequently, we adjusted the learning rate according to model size.

\subsection{Implementations}
\label{sec:imp}
Our implementations are based on huggingface transformers v4.42.1 \citep{transformers} using PyTorch v2.3.1 \citep{pytorch} and LlamaFactory \cite{llamafactory}.

For CasCoD \cite{improve}, we adhere to the optimal settings recommended by the authors, specifically setting $\alpha$ to 0.3. For KARD \cite{kard}, we employ the BM25 configuration \cite{bm25}, a sparse retrieval method based on word frequency, and retrieve three documents per question. Wikipedia serves as the external knowledge base for all datasets. For all retrievers used in NesyCD, including BM25, Contriever \cite{contriever}, and DPR \cite{DensePR}, we utilize the Pyserini\footnote{https://github.com/castorini/pyserini} library, which offers a reproducible information retrieval framework.

\subsection{Specialized Knowledge Collection}
\label{app_know}

For specialized knowledge collection, using 2-shot hand-written examples, the teacher model is configured with a temperature of 0.8 and a maximum length of 1024 tokens. It generates specialized knowledge corresponding to each incorrect example produced by the student SLMs. The prompt can be found in the Appendix \ref{prompt_know}.

\section{Additional Experiments and Findings}

\subsection{Why Choose Model Confidence?}
\label{confi}

We employ greedy decoding for $T = 16$ steps to calculate the confidence $\Delta_{\text{answer[: T]}}$. We posit that utilizing confidence scores to decide whether to retrieve information can strike a balance between performance and overhead. This approach potentially avoids the need for training a separate discriminator to ascertain the necessity of retrieval based on the question or explicitly asking whether knowledge is required \cite{self-knowledge}.

\section{Limitations}

\noindent \textbf{Method} We have demonstrated significant improvements in the performance of SLMs on complex reasoning tasks through NesyCD. However, it is important to acknowledge the limitations of our research. Methodologically, the effectiveness of our knowledge enhancement largely depends on the relevant knowledge retrieved from the external specialized knowledge base generated by the teacher LLM. While NesyCD enhances the complex reasoning performance of SLMs by utilizing a symbolic knowledge base to store additional knowledge, there remains a considerable gap between this and the gold standard knowledge. This suggests that the generalizability of the generated specialized knowledge can be further enhanced, thereby reducing the gap with the gold standard knowledge and allowing different models to be enhanced for various tasks.

\noindent \textbf{Retrieval} Adaptive retrieval remains a significant research challenge for current LLMs, including determining whether the LLM is confident and capable of solving problems through attention weights, hidden representations, classifiers, confidence, or prompts. Our work is limited to using confidence to simply judge whether additional knowledge is needed to assist reasoning. This approach achieves a balance between effectiveness and resource usage without additional training of smaller models, and a quantitative analysis is used to compare thresholds. In the future, better standards and methods can be defined to assess the capabilities of the LLM.

\noindent \textbf{Large Language Models} Regarding the experiments, given our limited computing and financial budgets, we chose GPT-3.5-Turbo as the teacher. Using GPT-4 would likely better verify the effectiveness of our method, NesyCD. Additionally, our aim to enhance the complex reasoning ability of SLMs restricted our choice to mainstream models smaller than 7B, such as Llama2, Llama3, and Qwen2, thereby excluding other excellent models like Phi3 and DeepSeek. However, exploring larger LMs such as 13B and 72B with NesyCD could be of great interest, presenting a promising direction for future research. Experimental results indicate that enhancing powerful models like Llama3-8B and Qwen2-7B surpasses GPT-3.5-Turbo and matches Llama3-70B.

\section{Extended Results}
\label{exre}
In Table \ref{ext}, we present the results of various models discussed in this paper, including LLaMA3-8B, QWen2-0.5B, 1.5B, and 7B, utilizing different baseline methods along with the outcomes of NesyCD.

\begin{table*}[t]
  \centering
  \renewcommand\arraystretch{1.05}
  \resizebox{\linewidth}{!}{
      \begin{tabular}{lcccccccc}
        \toprule
         \multirow{2}{*}{\textbf{Methods}} & \multicolumn{2}{c}{\textbf{In-Domain}}& \multicolumn{5}{c}{\textbf{Out-Of-Domain}} & \multirow{2}{*}{\textbf{Average}}\\
        \cmidrule(r){2-3}\cmidrule(r){4-8}
         & \textbf{BBH-test} & \textbf{GSM8K} & \textbf{BB-sub} & \textbf{AGIEval} & \textbf{GSM8K-PLUS} & \textbf{ARC-E}  & \textbf{ARC-C} &  \\
        \midrule
        \multicolumn{9}{l}{\textit{\# Closed-source model and Open-source models (Zero-shot-CoT)}}\\
        GPT-3.5-turbo (\textit{Teacher}) & 43.2 & 72.6 & 44.0 & 50.5 & 55.9 & 91.8 & 84.1 & 63.2 \\
        LLaMA-3-70B-Instruct & 62.6 & 89.2 & 51.0 & 66.3 & 72.9 & 97.6 & 93.2 & 76.1\\
        \midrule
        \multicolumn{9}{l}{\textit{\# LLaMA-3-8B based}}\\
        Zero-shot \cite{zeroshot}   & 18.2      & 2.8  & 27.4  & 29.7   & 2.2   & 50.8  & 50.0 &  25.9  \\
        Zero-shot-CoT \cite{zeroshotcot} & 26.5 & 6.6  & 23.5 & 32.2    & 3.7   & 68.1  & 55.5  & 30.9\\
        \hdashline
        Std-CoT \cite{stdcot}  & 79.4 & 61.6  & 40.5 & 41.3    & 45.6   & 83.2  & 71.9  & 60.5   \\
        MT-CoT \cite{mt-cot}   & 62.8 & 13.1  & 36.3 & 43.9    & 11.4    & 83.6  & 72.3  & 46.3   \\ 
        Step-by-step \cite{distilling}    & 64.0 & 11.5  & 38.8 & 43.7    & 9.0    & 84.3  & 74.6  & 46.6   \\
        KARD (BM25) \cite{kard}     & 81.4 & 64.3  & \textbf{43.1} & 43.4    & 48.6   & \textbf{85.6}  & \textbf{76.1}  & \textbf{63.2}   \\ 
        CasCoD \cite{improve}  & 32.1    & 59.1     & 18.1    & 23.6    &46.1   & 34.6     & 27.7  & 34.5   \\ 
        NesyCD (\textit{ours}) & \textbf{82.2} & \textbf{64.9}  & 42.2 & \textbf{44.1}    & \textbf{49.1}   & 84.7  & 75.4  & \textbf{63.2} \\
        \midrule
        \multicolumn{9}{l}{\textit{\# Qwen2-0.5B based}}\\
        Std-CoT \cite{stdcot}  & 65.8 & 26.7 & 29.6 & 25.6 & 17.1 & 43.6 & 32.0  & 34.3   \\
        MT-CoT \cite{mt-cot}   & 47.2 & 5.3 & 30.5 & 27.7 & 4.4 & 46.0 & 35.1  & 28.0   \\ 
        Step-by-step \cite{distilling}    & 44.2 & 5.2 & 28.9 & 26.2 & 3.1 & 41.8 & 36.2  & 26.5   \\
        KARD (BM25) \cite{kard}     & 66.3 & 30.9 & 31.7 & 23.9 & 18.2 & \textbf{48.9} & \textbf{37.2}  & 36.7   \\ 
        CasCoD \cite{improve}  & 37.6 & 27.7 & 20.0 & 15.6 & 17.6 & 21.5 & 14.8  & 22.1   \\ 
        NesyCD (\textit{ours}) & \textbf{68.7} & \textbf{32.2} & \textbf{31.8} & \textbf{28.4} & \textbf{19.5} & 46.8 & 36.7  & \textbf{37.7} \\
        \midrule
        \multicolumn{9}{l}{\textit{\# Qwen2-1.5B based}}\\
        Std-CoT \cite{stdcot}  & 68.2 & 52.7 & 35.7 & 34.0 & 37.3 & 69.3 & 56.4  & 50.5   \\
        MT-CoT \cite{mt-cot}   & 58.0 & 6.7 & 36.4 & 34.2 & 6.1 & 72.7 & 57.5  & 38.8   \\ 
        Step-by-step \cite{distilling}    & 48.4 & 5.8 & 32.8 & 34.4 & 6.1 & 72.1 & 57.6  & 36.7   \\
        KARD (BM25) \cite{kard}     & 72.2 & 55.4 & \textbf{37.4} & 31.2 & 39.4 & \textbf{74.0} & \textbf{62.2} & 53.1  \\ 
        CasCoD \cite{improve}  & 31.7 & 53.4 & 25.4 & 24.7 & 38.8 & 57.1 & 47.8  & 39.8   \\ 
        NesyCD (\textit{ours}) & \textbf{74.6} & \textbf{55.8} & \textbf{37.4} & \textbf{35.1} & \textbf{40.4} & 73.6 & 58.2  & \textbf{53.6} \\
        \midrule
        \multicolumn{9}{l}{\textit{\# Qwen2-7B based}}\\
        Std-CoT \cite{stdcot}  & 80.7 & 71.5 & 43.4 & 49.9 & 60.0 & 90.5 & 80.3 & 68.0   \\
        MT-CoT \cite{mt-cot}   & 70.0 & 15.2 & 42.6 & 49.4 & 12.1 & 90.9 & 80.2  & 51.5   \\ 
        Step-by-step \cite{distilling}    &  68.8 & 15.2 & 41.2 & 49.1 & 10.9 & 72.1 & 71.8  & 47.0   \\
        KARD (BM25) \cite{kard}     & 80.2 & 75.3 & 43.2 & 49.6 & 60.6 & \textbf{92.1} & \textbf{83.5} & 69.2  \\ 
        CasCoD \cite{improve}  & 35.7 & 72.3 & 23.8 & 37.4 & 60.6 & 70.1 & 63.1  & 51.9   \\ 
        NesyCD (\textit{ours}) & \textbf{80.9} & \textbf{76.3} & \textbf{43.9} & \textbf{49.9} & \textbf{60.7} & 91.9 & 82.9  & \textbf{69.5} \\
        \bottomrule
      \end{tabular}
      }
    \caption{Performance (\%) of LLaMA3-8B \cite{llama3} and Qwen2-0.5B, 1.5B and 7B \cite{Qwen2TR} with different methods across seven selected datasets. \textbf{Bold} indicates the best in each setting.}
    \label{ext}
\end{table*}

\section{Case Study}
\label{case}

We present two cases from Tables \ref{tab:gsm} and \ref{tab:bbh} to compare the Chains of Thought (CoTs) generated by NesyCD, the teacher large language model (LLM), and the standard CoTs distillation method (Std-CoT). We use \textcolor{green}{\ding{51}} and \textcolor{red}{\ding{55}} to indicate the correctness of the CoT. 

Table \ref{tab:gsm} shows that the Std-CoT's response is confused and fails to comprehend the question accurately. Although it has a rough idea, its rationale is entirely incorrect as it struggles to emulate the rationale of the teacher LLM. Table \ref{tab:bbh} illustrates that, without background knowledge and summary learning, even the teacher LLM may make errors in complex reasoning tasks such as path planning, owing to its poor understanding of spatial imagination and coordinate positions. However, with specialized knowledge, NesyCD can accurately grasp position information and orientation logic, enabling it to generate a correct, concise, and rich rationale.

\begin{table*}[htbp]
\centering
\renewcommand\arraystretch{1.15}
\resizebox{\linewidth}{!}{
\begin{tabular}{lp{8cm}p{8cm}}
\hline
\multicolumn{3}{l}{\makecell[l]{Question: Pauline visits her favorite local museum 3.0 times a year. The cost of one visit is \$2.00. After 5.0 years, the cost of one\\ visit has increased by 1.5 times, but Pauline decided not to give up any visit and continued to go to the museum for 3/1 more years.\\How much did Pauline spend on all visits to the museum in dollars?}}
\\
\hline
 & \textbf{Specialized Knowledge} & \textbf{Output} \\
\hline
\textbf{Teacher} & - & Pauline visited the local museum 3 times a year, so she paid 3 * 2 = 6 every year on visits. That means during the first 5 years Pauline spend 5 * 6 = 30 on tickets. After 5 years the cost of one visit increased by 150/100 * 2 = 3. That means the new price of the ticket was 2 + 3 = 5. So for per year Pauline paid 3 * 5 = 15. During the last 3 years, she paid in total 3 * 15 = 45. In total Pauline spend 30 + 45 = 75 on visits at the museum. Therefore, the answer is 75. ~~~ ~~~ ~~~\textcolor{green}{\ding{51}}\\
\hline
\textbf{Std-CoT} & - & 3.0 times a year means Pauline goes to the museum 3.0 * 5.0 = 15.0 times in 5.0 years. The cost of one visit is \$2.00, but in the last 5.0 years, it has increased by 1.5 times, so \textcolor{red}{the new cost is \$2.00 * 1.5 = \$3.00}. So, \textcolor{red}{Pauline spent 15.0 * 3.0 = \$45.00 on all visits to the museum in dollars}. Therefore, the answer is 45.  ~~~ ~~~ ~~~\textcolor{red}{\ding{55}}\\
\hline
\textbf{NesyCD} & Learning Summary: 1. Careful Reading and Understanding: It is crucial to read the question carefully to grasp the scenario and requirements, such as \textcolor{brown}{the frequency of visits and changes in ticket prices over time}. 2. Accurate Mathematical Operations: Ensure that \textcolor{brown}{calculations reflect the progression of events} accurately, such as calculating the total cost for each period based on the given conditions. (...) Supplementary Knowledge: 1. Unit Conversion: Understand how to convert between different units, such as months to years, and ensure that the conversions align with the context of the question. 2. Cost Calculations: Practice calculating total costs based on given conditions, including \textcolor{brown}{fixed fees and variable costs that may change over time or with specific actions}. (...) & 3 years later, the cost of one visit is \$2 * 1.5 = 3.00 more than the previous cost. One visit now costs \$3.00 + \$2.00 = \$5.00. Pauline visited the museum 3 * 5 = 15 times in 3 years. In total, Pauline spent \$5 * 15 = \$75.00 on all visits in dollars. Therefore, the answer is 75.  ~~~ ~~~ ~~~\textcolor{green}{\ding{51}}\\
\hline
\end{tabular}}
\caption{A integer-decimal-fraction conversion case from GSM8K+.}
\label{tab:gsm}
\end{table*}

\begin{table*}[htbp]
\centering
\renewcommand\arraystretch{1.15}
\resizebox{\linewidth}{!}{
\begin{tabular}{lp{8cm}p{8cm}}
\hline
\multicolumn{3}{l}{\makecell[l]{Question: If you follow these instructions, do you return to the starting point? Take 9 steps. Turn left. Turn left. Take 9 steps.\\Options:\\- Yes\\- No}}
\\
\hline
 & \textbf{Specialized Knowledge} & \textbf{Output} \\
\hline
\textbf{Teacher} & - & We start at the origin (0, 0), facing the positive y-axis. (1) Take 9 steps: (0, 9), facing the positive y-axis. (2) \textcolor{red}{Turn left: (9, 0), facing the negative x-axis. (3) Turn left: (9, 0), facing the positive x-axis. (4) Take 9 steps: (18, 0), facing the positive x-axis.} Since (18, 0) is not (0, 0), we are not where we started. Therefore, the answer is No. ~~~ ~~~ ~~~\textcolor{red}{\ding{55}}\\
\hline
\textbf{Std-CoT} & - & We start at the origin (0, 0), facing the positive y-axis. (1) Take 9 steps: (0, 9), facing the positive y-axis. (2) Turn left: (0, 9), facing the negative x-axis. (3) \textcolor{red}{Turn left: (0, 9), facing the positive x-axis. (4) Take 9 steps: (9, 9), facing the positive x-axis}. Since (9, 9) is not (0, 0), we are not where we started. Therefore, the answer is No.  ~~~ ~~~ ~~~\textcolor{red}{\ding{55}}\\
\hline
\textbf{NesyCD} & Learning Summary: 1. Sequential Execution of Instructions: \textcolor{brown}{Follow the given steps in the order provided}, ensuring that each instruction is carried out accurately based on the previous position. 2. Spatial Awareness: \textcolor{brown}{Maintain awareness of spatial coordinates and directions} while moving according to the instructions to accurately determine the final position. 3. Consistency in Tracking: \textcolor{brown}{Keep track of the current position after each step} to avoid discrepancies or errors in determining the final destination. (...)  Supplementary Knowledge: 1. Spatial Coordinates: \textcolor{brown}{Understand the concept of coordinates} and how movements in different directions affect the position on a grid or map. 2. Directional Instructions: \textcolor{brown}{Recognize and interpret directional commands} such as turning around, moving forward, or changing positions to navigate a path effectively. (...) & 1. Take 9 steps: (0), 9. 2. Turn left: (0), 9, left. 3. Turn left: (0), 9, left, left. 4. Take 9 steps: (0), 9, left, left, 9. Now, we are where we started with a total distance traveled of 18 steps. Therefore, the answer is Yes.  ~~~ ~~~ ~~~\textcolor{green}{\ding{51}}\\
\hline
\end{tabular}}
\caption{A navigate case from BBH-test.}
\label{tab:bbh}
\end{table*}

\section{Instruction Details}

\subsection{Prompt for Generating CoTs}
\label{prompt_cot}

We use the prompt template shown below to call the teacher model to generate the CoTs for the training datasets.

\newpage 
    ~ 
  \newpage   
\begin{tcolorbox}
    [width=\textwidth, title = {Generate CoTs}]
    
    You are an expert assistant teacher. The following are tasks about \{Task\_Name\}. \{Task Description\}. Explain your reasoning first and your response should conclude with the format ``Therefore, the answer is".

    ~ 
    
    Question: \{QUESTION\}
    
    Answer: Let’s think step by step.
\end{tcolorbox}

\subsection{Prompt for Specialized Knowledge Collection}
\label{prompt_know}

\noindent \textbf{Generate Learning Summary} only prompts LLMs to analyze the SLM's errors and generate the specialized knowledge of learning summary.

\newpage

~

\begin{tcolorbox}[width=\textwidth, title = {Generate Learning Summary}]

As an excellent educational teacher, your goal is to help students enhance their question-solving abilities by analyzing incorrect solution processes and answers.

Based on an understanding and explanation of the question, along with relevant background knowledge, fundamental concepts, and empirical conclusions, please generate a learning summary in a numbered list format that will help students complete the same task in the future.

    ~ 
    
\#\#\# Requirements:

1. Learning summary should outline the thought processes and precautions for addressing student mistakes, including, but not limited to, question comprehension, thought steps and mathematical calculations. It should also provide a summative experience to help students solve similar questions in the future.

2. Ensure that the content is understandable and usable by students, while also being concise and effective.

3. If the student's answer does not contain the format \"Therefore, the answer is\" or if the repeated output of certain characters indicates that the student cannot solve the question.

4. The obtained learning summary should be general and generalized, not aimed at specific questions.

5. Keep these requirements in mind while generating the learning summary and supplementary knowledge.
  
    ~ 
    
\#\#\# Return Format:

Return in the following format:

Learning Summary: [Learning Summary]
  
    ~ 
    
The student was given the following question: \{QUESTION\}

The student’s wrong response is: \{WRONG RESPONSE\}

Please follow the requirements and provide the learning summary.
\end{tcolorbox}

\noindent \textbf{Generate Learning Summary and Supplementary Knowledge} prompts LLMs to analyze the SLM's errors and generate the specialized knowledge of learning summary and Supplementary Knowledge, providing additional relevant background knowledge to further assist SLMs in solving similar complex reasoning tasks in the future.

\newpage
\newpage
\newpage
\newpage

    ~ 
    
    ~ 
    
    ~ 
    
    ~ 
        ~ 
    
    ~ 
    
    ~ 
    
    ~ 
        ~ 
    
    ~ 
    
    ~ 
      ~ 
    
    ~ 
    
    ~ 
    
    ~ 
        ~ 
    
    ~ 
    
    ~ 
    
    ~ 
        ~ 
    
    ~ 
    
    ~ 
      ~ 
    
    ~ 
    
    ~ 
    
    ~ 
        ~ 
    
    ~ 
    
    ~ 
    
    ~ 
        ~ 
    
    ~ 
    
    ~ 
      ~ 
    
    ~ 
    
    ~ 
    
    ~ 
        ~ 
    
    ~ 
    
    ~ 
    
    ~ 
        ~ 
    
    ~ 
    
    ~ 
 
\begin{tcolorbox}[width=\textwidth, title = {Generate Learning Summary and Supplementary Knowledge}]
As an excellent educational teacher, your goal is to help students enhance their question-solving abilities by analyzing incorrect solution processes and answers. 
    
You should generate targeted, detailed thought processes and relevant background knowledge for solving similar questions in the future.

Your role involves creating learning summaries and supplementary knowledge, specifically identifying the steps needed to solve the question and providing additional general knowledge in the supplementary knowledge.

    ~ 
    
\#\#\# Requirements:

1. Learning summary should outline the thought processes and precautions for addressing student mistakes, including, but is not limited to, question comprehension, thought steps and mathematical calculations. It should also provide a summative experience to help students solve similar questions in the future.

2. Supplementary knowledge should include a list of essential background information that students need to solve the question. This should encompass but is not limited to, mathematical formulas, definitions, relevant world knowledge, and specific techniques.

3. Ensure that the content is understandable and usable by students, while also being concise and effective.

4. If the student's answer does not contain the format \"Therefore, the answer is\" or if the repeated output of certain characters indicates that the student cannot solve the question.

5. The obtained learning summary should be general and generalized, not aimed at specific questions, and the supplementary knowledge should also be general knowledge of the question without involving specific analysis.

6. Keep these requirements in mind while generating the learning summary and supplementary knowledge.
  
    ~ 
    
\#\#\# Return Format:

Return in the following format:

Learning Summary: [Learning Summary]

Supplementary Knowledge: [Supplementary Knowledge]
  
    ~ 
    
The student was given the following question: \{QUESTION\}

The student’s wrong response is: \{WRONG RESPONSE\}

Please follow the requirements and provide the learning summary and supplementary knowledge.
\label{full}
\end{tcolorbox}

\end{document}